\documentclass[letterpaper, 10 pt, conference]{ieeeconf}  % Comment this line out
\IEEEoverridecommandlockouts                              % This command is only
\usepackage[utf8]{inputenc}
\usepackage[T1]{fontenc}

\usepackage[table]{xcolor}
\definecolor{rblue}{rgb}{0,0.5,1}
\definecolor{awesome}{rgb}{1.0, 0.13, 0.32}
\definecolor{hollywoodcerise}{rgb}{0.96, 0.0, 0.63}
\definecolor{lasallegreen}{rgb}{0.03, 0.47, 0.19}
\definecolor{hanpurple}{rgb}{0.32, 0.09, 0.98}
\definecolor{green(pigment)}{rgb}{0.0, 0.65, 0.31}

\makeatletter
\let\NAT@parse\undefined
\makeatother
\usepackage[pagebackref=false, breaklinks=true, colorlinks, bookmarks=false]{hyperref}
\hypersetup{colorlinks,linkcolor={red},citecolor={hanpurple},urlcolor={magenta}}  

\usepackage{amsmath}
\usepackage{amssymb}
\usepackage{caption}
\usepackage{graphicx}
\usepackage{booktabs}
\usepackage{multirow}
\usepackage{placeins}

\title{\LARGE \bf
OmniMimic: Dynamics-completed Motion Augmentation for Multi-style Omnidirectional Quadruped Locomotion
}

\author{Sheng Wu$^{1,*}$, Guoqiang Zhao$^{1,*}$, Zhe Yang$^{1,*}$, Fei Teng$^{1,*}$, Zhikun Zhou$^{1}$, Yanlin Yang$^{2}$, Zheng Fang$^{2}$,\\Hong Zheng$^{2}$, Yaonan Wang$^{1,3}$, and Kailun Yang$^{1,3,\dag}$%
\thanks{This work was supported in part by the National Natural Science Foundation of China (Grant No. 62473139 and No. 62388101), in part by the Hunan Provincial Research and Development Project (Grant No. 2025QK3019), in part by the State Key Laboratory of Autonomous Intelligent Unmanned Systems (the opening project number ZZKF2025-2-10), and in part by China Mobile Hunan Company Limited and China Mobile Communications Group Co., Ltd., and was conducted under the project ``Research on Reinforcement Learning Algorithm for Quadruped Bionic Locomotion with Emotional Expression''.}
\thanks{$^{1}$The authors are with the School of Artificial Intelligence and Robotics, Hunan University, China (email: kailun.yang@hnu.edu.cn).}%
\thanks{$^{2}$The authors are with China Mobile Group Hunan Company Ltd., China.}
\thanks{$^{3}$The authors are also with the National Engineering Research Center of Robot Visual Perception and Control Technology, Hunan University, China.}%
\thanks{$^{*}$These authors contributed equally to this work and share first authorship.}
\thanks{$^{\dag}$Corresponding author: Kailun Yang.}
}

\let\oldtwocolumn\twocolumn
\renewcommand\twocolumn[1][]{%
    \oldtwocolumn[{#1}{
    \begin{center}
    \vskip -3ex
        \centering
        \includegraphics[width=0.99\textwidth]{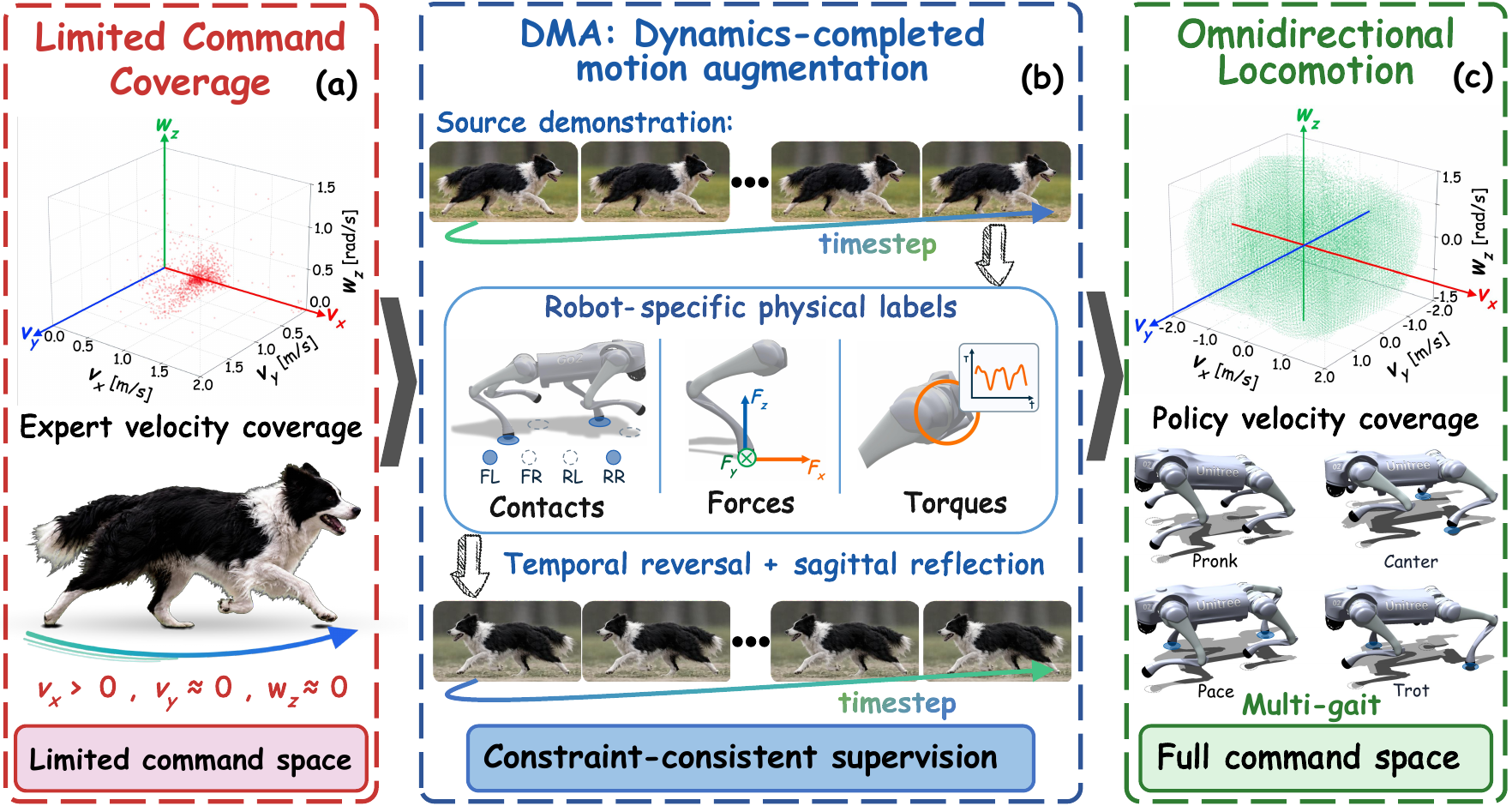}
        \vskip -1ex
        \captionof{figure}{Overview of OmniMimic.
        (a) Canine demonstrations provide limited coverage of the target velocity-command space.
        (b) Dynamics-completed Motion Augmentation (DMA) constructs robot-specific physical supervision beyond the demonstrated motions.
        (c) The resulting policy achieves multi-gait omnidirectional locomotion over the full command space.
        }
        \label{fig:overview}
    \end{center}
    }]
}

\begin{document}

\maketitle
\thispagestyle{empty}
\pagestyle{empty}

\begin{abstract}
Animal demonstrations provide quadruped robots with natural and distinctive gait styles that are difficult to specify through hand-crafted rewards. However, their narrow directional coverage leaves little style-consistent supervision for backward, lateral, and turning commands. We present OmniMimic, a training framework that turns directionally limited animal demonstrations into a single multi-gait policy over target per-axis velocity ranges. OmniMimic first combines temporal reversal, constrained dynamics completion, and sagittal reflection to construct robot-specific kinematic and physical supervision beyond the observed directions. It then expands commands progressively from the demonstrated velocity distribution toward the target per-axis bounds, and uses a shared actor with soft-gated, gait-specialized residual experts to balance reusable locomotion skills with gait-specific corrections. Across four gaits in simulation, OmniMimic reduces mean foot-position RMSE at forward and backward reference velocities by $12.9\%$ and velocity-tracking RMSE on a uniform Cartesian command grid by $63.1\%$, compared with the matched APEX baseline. The project page is at~\href{https://OmniMimic.github.io}{https://OmniMimic.github.io}.
\end{abstract}

\section{Introduction}
\label{sec:introduction}

Quadruped robots are moving from controlled demonstrations toward applications such as companionship~\cite{tamura2004entertainment}, interactive entertainment~\cite{han2024lifelike}, and industrial inspection~\cite{gehring2021anymal}.
These settings call for locomotion that is not only stable, but also expressive across gait styles and responsive to omnidirectional velocity commands.
Animal-motion imitation provides an effective way to acquire natural gait styles because demonstrations contain rich coordination patterns that are difficult to encode with hand-crafted rewards~\cite{peng2020learning,escontrela2022adversarial}.
However, natural canine locomotion is dominated by forward progression~\cite{vilensky2000quadrupeds}, whereas lateral and backward motions are comparatively rare~\cite{mccauley2018therapeutic,zink2025locomotion}.
Consequently, the resulting motion library covers only a limited region of the target omnidirectional command space, as illustrated in Fig.~\ref{fig:overview}(a).

\begin{figure}[!t]
    \centering
    \includegraphics[width=\columnwidth]{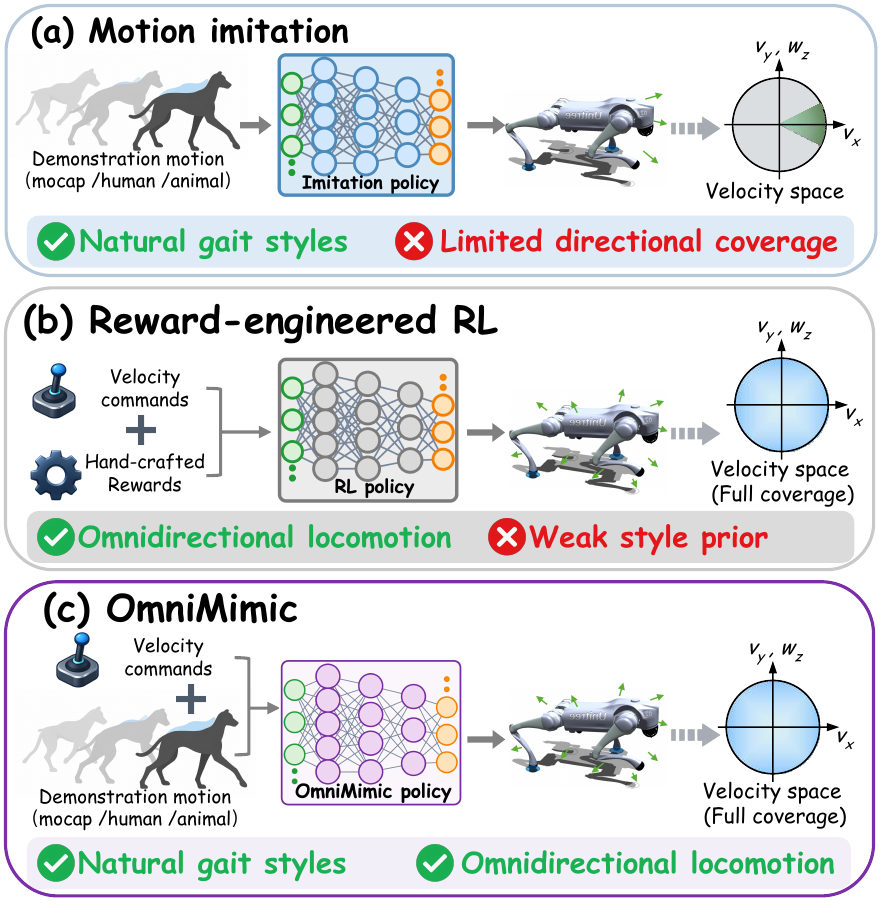}
    \vskip -1ex
    \caption{
    Comparison of quadruped locomotion learning paradigms.
    Directionally limited demonstrations provide sparse style supervision away from recorded motions; reward-engineered RL uses explicit objectives to shape gait behavior across broad commands. OmniMimic combines demonstration-derived gait priors with omnidirectional command following.
    }
    \label{fig:motivation}
    \vskip-4ex
\end{figure}

This setting highlights the challenge of combining demonstration-derived gait styles with broad command following (Fig.~\ref{fig:motivation}).
Motion imitation provides natural gait priors~\cite{peng2020learning,escontrela2022adversarial,peng2018deepmimic,sood2025apex}, but directionally limited demonstrations supply sparse style supervision for commands far from the recorded motions. Reward-engineered reinforcement learning can achieve broad command coverage~\cite{hwangbo2019learning,rudin2022learning,margolis2023walk}, with timing, posture, and foot-motion objectives used to shape gait behavior~\cite{margolis2023walk,kim2025learning}.
Our task is to learn one policy that retains the demonstrated gait styles while following commands to move backward, move sideways, and turn, including at velocities absent from the demonstrations.
\mbox{OmniMimic} combines physical supervision for reversed and reflected references with progressive command expansion and a shared-residual actor to learn multi-gait omnidirectional control, as shown in Fig.~\ref{fig:overview}(b)--(c).

Three difficulties make this extension nontrivial. First, reversing a pose sequence does not determine the contact forces and joint torques required by the target robot, and temporal reversal alone does not provide the left--right transformation needed to pair leftward and rightward motion or clockwise and counterclockwise turns. Second, exposing the policy to the target per-axis ranges from the start creates an abrupt distribution shift from the demonstrated velocities and can destabilize style learning~\cite{rudin2022learning}.
Third, a single fully shared actor must reconcile reusable locomotion structure with gait-specific coordination, which can cause distinct gaits to collapse toward a common motion.
We address these challenges through physical supervision for augmented references, progressive command expansion, and shared policy capacity with gait-specialized residuals.

We present \textbf{OmniMimic}, a training framework for learning one command-conditioned policy from four demonstrated gait styles over target omnidirectional velocity ranges. Reference motions and physical supervision are used only during training; the deployed actor receives proprioception, user commands, and a gait code.
The matched four-gait evaluation shows $12.9\%$ lower foot-position RMSE at forward/backward reference velocities and $63.1\%$ lower full-grid velocity-tracking RMSE than APEX (Table~\ref{tab:independent_baselines}).
\mbox{Our contributions are:}

\begin{itemize}
    \begin{samepage}
    \item \textbf{Dynamics-completed Motion Augmentation (DMA).} We construct robot-specific physical targets for time-reversed and sagittally reflected references, and incorporate them through confidence-weighted torque guidance and event-aware rewards.
    \par\end{samepage}

    \item \textbf{Demonstration-guided Omnidirectional Command Expansion (DOCE).} We progressively move from frame-matched reference velocities to independently sampled target commands on all three axes, gradually increasing the distance from demonstrated velocities.

    \item \textbf{Shared-residual Multi-gait Policy (SMP).} We pair a shared actor with bounded residual experts mixed by a gait-conditioned soft gate, providing shared control capacity and gait-dependent corrections within one policy.
\end{itemize}

\section{Related Work}
\label{sec:related_work}

\subsection{Demonstration-based Motion Imitation}

Motion imitation transfers natural gait styles from demonstrations to legged robots~\cite{peng2020learning,peng2018deepmimic,peng2021amp,mirza2025imitation}. Reference-tracking methods follow time-indexed trajectories~\cite{peng2018deepmimic,peng2020learning}, whereas AMP matches motion distributions without frame-wise tracking~\cite{peng2021amp}. Multi-style motion priors~\cite{vollenweider2023advanced,mu2026smp,yang2025gai} and reusable adversarial skill embeddings~\cite{peng2022ase} support diverse, controllable behaviors. DecAP~\cite{sood2024decap} and APEX~\cite{sood2025apex} use decaying action- or torque-level priors to preserve reference characteristics while allowing task exploration.
Hierarchical residual learning adapts flat-terrain animal-motion priors to rough terrain~\cite{zhang2025motionpriors}, while Uni-Mo generates robot-motion videos and lifts them into 3D references for tracking-policy training~\cite{liu2026unleashing}.
Rather than generating new motion categories, OmniMimic extends the directional and velocity coverage of given animal gait demonstrations by constructing style-consistent kinematic and physical supervision for unobserved commands.
\begin{figure*}[t]
    \centering
    \includegraphics[width=\textwidth]{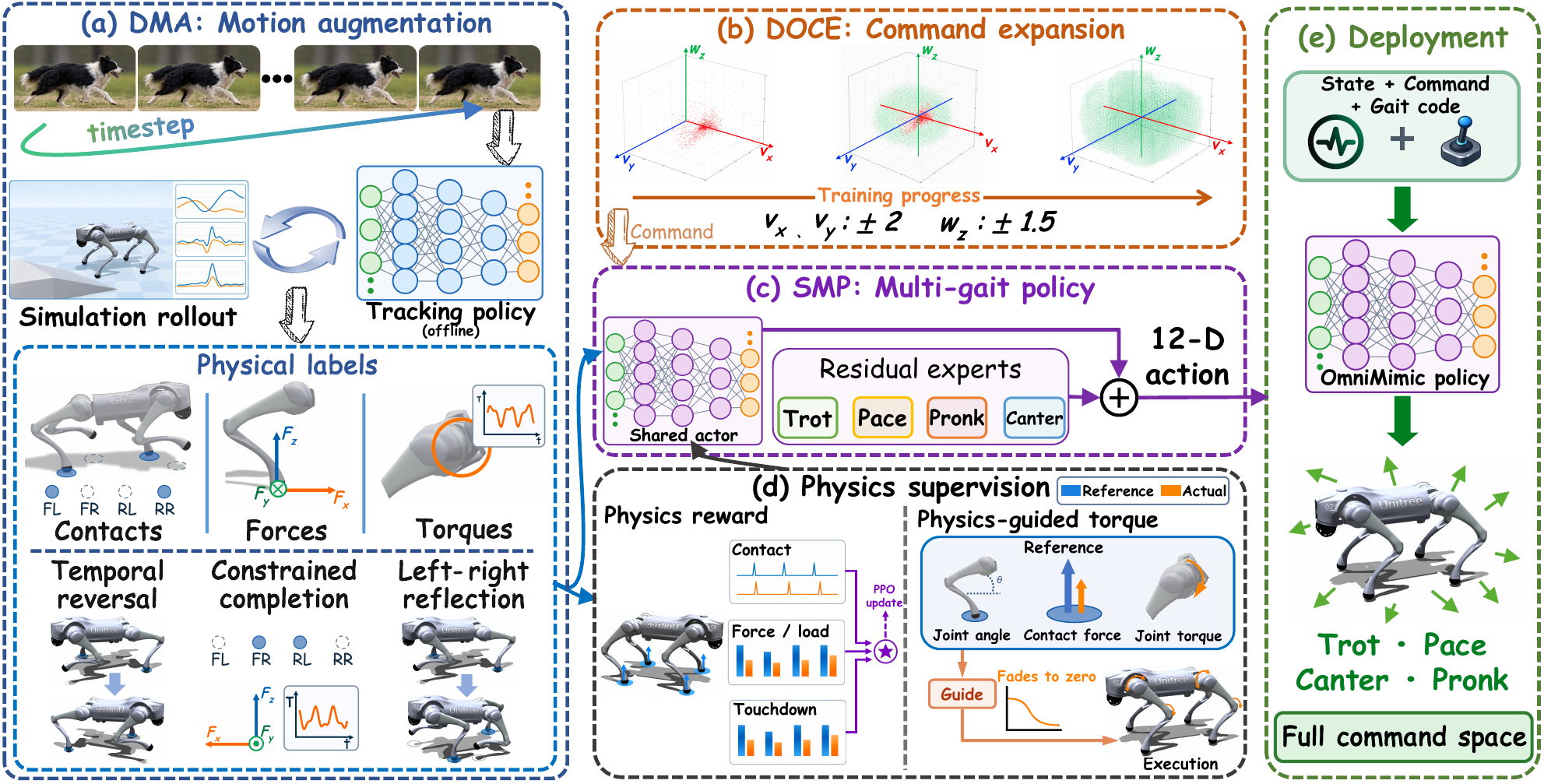}
    \vskip-1ex
    \caption{
    Overview of OmniMimic.
    (a) Dynamics-completed Motion Augmentation (DMA) produces reversed and mirrored references with physical labels.
    (b) Demonstration-guided Omnidirectional Command Expansion (DOCE) expands per-axis command ranges.
    (c) Shared-residual Multi-gait Policy (SMP) adds softly gated residuals to a shared action.
    (d) Training uses physical guidance and event-aware rewards.
    (e) Actor-only deployment uses proprioception, commands, and gait code.
    }
    \label{fig:method_overview}
    %\vspace{-20pt}
    \vskip-4ex
\end{figure*}

\subsection{Omnidirectional and Multi-gait Locomotion}

Reward-driven reinforcement learning achieves robust velocity tracking by training directly over broad command distributions~\cite{hwangbo2019learning,rudin2022learning}.
Related advances include online adaptation~\cite{kumar2021rma}, perception-aware omnidirectional skills~\cite{li2025move}, and direct on-robot omnidirectional learning~\cite{bohlinger2025gait}. Adaptive velocity-command curricula enable high-speed locomotion~\cite{margolis2022rapid}.
``Walk These Ways'' exposes gait, swing, posture, and speed parameters~\cite{margolis2023walk}; Gaitor learns a unified gait representation~\cite{mitchell2024gaitor}; and MELA combines expert networks through learned gating~\cite{yang2020mela}. Biomechanics-inspired controllers select gaits online~\cite{humphreys2025learning}, while style rewards explicitly encode timing, clearance, or posture~\cite{kim2025learning}.
OmniMimic instead targets style-consistent command expansion from limited animal demonstrations. Its curriculum begins at matched reference velocities, and its shared actor combines bounded residual corrections rather than synthesizing a policy from complete expert networks. The final policy jointly learns gait-conditioned locomotion and command tracking over the target velocity ranges.

\subsection{Physics-aware Motion Processing}

Physics-aware motion processing aligns demonstrations with target-robot dynamics. STMR~\cite{yoon2025stmr}, ``Walk Like Dogs''~\cite{kang2026walkdogs}, and ReActor~\cite{muller2026reactor} improve feasibility through constrained retargeting, kino-dynamic optimization, or joint reference-policy optimization. ADP~\cite{lee2026adp} and APT-RL~\cite{kang2026aptrl} incorporate dynamics priors or torque information into learning. Reverse playback can also provide an exploration prior for an inverse skill; Multi-AMP, for example, uses a reversed stand-up motion to help discover sitting down~\cite{vollenweider2023advanced}. Morphological symmetries provide consistent state and dynamics transformations~\cite{ordonez2025morphological}, supporting symmetry-aware augmentation and actor-critic architectures~\cite{su2024symmetry}. OmniMimic complements these approaches by collecting robot-specific physical labels from forward motions tracked in the target-robot simulator, completing force and torque targets for reversed motions, and consistently reflecting kinematic and physical quantities. The resulting labels provide training-only supervision for multi-gait command expansion.

\section{OmniMimic Framework}
\label{sec:method}

\textbf{OmniMimic} learns a single multi-gait policy from animal demonstrations with limited directional coverage.
As shown in Fig.~\ref{fig:method_overview}, it combines Dynamics-completed Motion Augmentation (DMA), Demonstration-guided Omnidirectional Command Expansion (DOCE), and a Shared-residual Multi-gait Policy (SMP) to augment physical supervision, broaden per-axis command ranges, and share control across gaits.
Numerical settings are summarized in Table~\ref{tab:training_parameters}.

\subsection{Dynamics-completed Motion Augmentation (DMA)}

\emph{\textbf{Obtaining physical labels.}}
For each source motion, we train a reference-conditioned tracking policy in simulation for offline physical-label collection, following~\cite{sood2025apex}.
It receives robot state, reference-motion features, and the normalized reference-frame index $\ell/(N-1)$, with $\ell$ the zero-based index in an $N$-frame sequence; episodes start from reference poses and velocities.
We record joint states, applied torques, foot contacts, and forces by reference frame, using the Go2 model and Pinocchio~\cite{carpentier2019pinocchio} for dynamics terms and foot Jacobians.
These tracking policies are distinct from the final command-conditioned actor: they are neither deployed nor used to initialize it. The final actor is trained from scratch without reference-motion or reference-frame-index inputs.

\emph{\textbf{Completing time-reversed supervision.}}
We reverse frame order, negate joint and base velocities, and re-anchor root translation to construct backward references.
Because recorded forward forces and torques need not satisfy the reversed dynamics, we fix each reversed generalized configuration and velocity $(\mathbf q^{\rm rev},\mathbf v^{\rm rev})$ and solve
\begin{equation}
\begin{aligned}
\min_{\dot{\mathbf v},\mathbf f,\boldsymbol\tau}\quad
& \|\dot{\mathbf v}-\dot{\mathbf v}^{\rm rev}\|_{W_{\dot v}}^{2}
+\|\mathbf f-\mathbf f^{\rm seed}\|_{W_f}^{2} \\
& +\|\boldsymbol\tau-\boldsymbol\tau^{\rm seed}\|_{W_\tau}^{2}
+\mathcal R_{\rm smooth}(\mathbf f,\boldsymbol\tau) \\
\text{s.t.}\quad
& M(\mathbf q^{\rm rev})\dot{\mathbf v}
+C(\mathbf q^{\rm rev},\mathbf v^{\rm rev})\mathbf v^{\rm rev}
+\mathbf g(\mathbf q^{\rm rev}) \\
& \qquad =S^\top\boldsymbol\tau
+J_c^\top(\mathbf q^{\rm rev})\mathbf f, \\
& -\boldsymbol\tau_{\max}
\leq\boldsymbol\tau
\leq\boldsymbol\tau_{\max}, \\
& 0\leq f_{j,z}\leq f_{z,\max},
\qquad j\in\mathcal S_{\rm stance}, \\
& |f_{j,x}|,\ |f_{j,y}|
\leq \mu f_{j,z},
\qquad j\in\mathcal S_{\rm stance}, \\
& \mathbf f_j=\mathbf 0,
\qquad j\in\mathcal S_{\rm swing}.
\end{aligned}
\label{eq:dynamics_completion}
\end{equation}
Here $\dot{\mathbf v}$, $\mathbf f$, and $\boldsymbol\tau$ are generalized acceleration, stacked foot forces, and joint torques.
$M$, $C\mathbf v$, and $\mathbf g$ give the mass, Coriolis/centrifugal, and gravity terms; $J_c$ is the contact Jacobian and $S$ selects actuated joints.
The diagonal weights $W_{\dot v},W_f,W_\tau$ penalize squared deviations from reversed kinematic acceleration $\dot{\mathbf v}^{\rm rev}$ and temporally reordered forward measurements $(\mathbf f^{\rm seed},\boldsymbol\tau^{\rm seed})$; $\mathcal R_{\rm smooth}$ penalizes squared force/torque changes from the preceding completed frame.
The forward measurements serve as optimization priors rather than ground truth for reversed motion.
For foot $j$, $\mathcal S_{\rm stance}$ and $\mathcal S_{\rm swing}$ identify stance and swing; $\boldsymbol\tau_{\max}$ and $f_{z,\max}$ bound torques and normal force, and $\mu$ is the friction coefficient in the friction-pyramid approximation.
This frame-wise solve supplies supervision labels, not a separately integrated trajectory.

Forward-label validity requires finite forces, torques, and Jacobians; confidence decays exponentially with joint-position tracking RMSE and is reduced at contact transitions.
Completed labels require finite forces and torques and dynamics/constraint residuals within numerical tolerances.
For completed labels, confidence is a clipped product of exponential penalties on base and joint acceleration corrections and contact-event factors.
These factors reduce confidence at contact transitions, when touchdown impulses exceed the forward-rollout envelope, or when impulse reconstruction fails.
Forces and contact timing yield vertical-load shares and touchdown impulses.
Integer source-frame lookup transfers labels to four-gait references using forward or reversed indices, without interpolation or further dynamics solves.

\emph{\textbf{Sagittal-plane reflection.}}
To add left--right counterparts, we reflect each forward or reversed reference across the sagittal plane: exchange FL/FR and RL/RR, transform joint coordinates, and negate lateral velocity and yaw rate.
The same transformation is applied to forces, load shares, contact events, torques, and Jacobians, pairing mirrored commands with consistent motion and physical labels.

\emph{\textbf{Physics-guided policy learning.}}
The label bank provides torque guidance and event-aware rewards during training (Fig.~\ref{fig:method_overview}(d)).
A \emph{physical torque residual} supplements the reference-position-based guide in~\cite{sood2025apex}:
\begin{equation}
\boldsymbol\tau^{\rm guide}_t=d_t\!\left[
\boldsymbol\tau^{\rm kin}_t+
\alpha(i)g_t\,\operatorname{clip}_{\gamma_\tau\boldsymbol\tau_{\max}}
\!\left(\Delta\boldsymbol\tau^{\rm phy}_t\right)\right].
\label{eq:physics_guide}
\end{equation}
Here $t$ and $i$ index control steps and training iterations.
The kinematic correction $\boldsymbol\tau^{\rm kin}_t=K_p(\mathbf q_{{\rm J},t}^{\rm ref}-\mathbf q_{{\rm J},t})$ uses proportional gain $K_p$ and reference/current actuated-joint positions; $d_t$ decays exponentially with accumulated training control steps $n_t$.
Only the physical residual is clipped to a fraction $\gamma_\tau$ of torque limits and modulated by the half-cosine decay $\alpha(i)$ and gate $g_t$.
This gate multiplies command compatibility $g_t^{\rm cmd}$ by the larger of the weighted reference stable-support and near-touchdown indicators.
The touchdown indicator requires reference contact, keeping the gate zero during reference flight.
Command compatibility is the exponential of negative mean squared command--reference error after per-axis tolerance normalization.
The residual is
\begin{equation}
\begin{aligned}
\Delta\boldsymbol\tau^{\rm phy}_t
={}&w_v K_d\dot{\mathbf q}^{\rm ref}_{{\rm J},t} \\
&+b_t\!\left[w_\tau c_t\boldsymbol\tau^{\rm comp}_t
+w_s(1-c_t)\boldsymbol\tau^{\rm seed}_t\right] \\
&+w_f\sum_{j=1}^{4}\eta_{j,t}J_{j,t}^{\top}
\!\left(\mathbf f_{j,t}^{\rm ref}-\mathbf f_{j,t}^{\rm actual}\right).
\end{aligned}
\label{eq:physics_residual}
\end{equation}
The weights $(w_v,w_\tau,w_s,w_f)$ scale reference-velocity feedforward, completed torque, seed torque, and force feedback; $K_d$ is the derivative gain and $\dot{\mathbf q}^{\rm ref}_{{\rm J},t}$ is the reference joint velocity.
The mask $b_t\in\{0,1\}$ is one for valid labels; confidence $c_t\in[0,1]$ follows the checks above.
The per-foot factor $\eta_{j,t}$ is the product of label validity $b_t$, confidence $c_t$, and the reference-contact and non-transition masks; it gates force feedback to reliable, non-transition reference contacts.
Foot $j$'s reference/measured forces and actuated-joint Jacobian $J_{j,t}$ use heading-aligned coordinates.
The completed and seed torques provide a feedforward prior, with lower confidence shifting their mixture toward the measured seed. In contrast, the force term feeds back the reference--measured force error and vanishes when the forces match; it does not add the reference contact wrench a second time.
Guidance is added to the actor-driven PD torque before actuator limiting.

An \emph{event-aware physics reward} evaluates physical matching only during relevant gait events:
\begin{equation}
r_t^{\rm phy}=\lambda\,\rho(i)\,g_t^{\rm cmd}\,\chi_t
\frac{\sum_{k\in\mathcal K}\beta_k(i)m_{k,t}R_{k,t}}
{\max\!\left(\sum_{k\in\mathcal K}\beta_k(i)m_{k,t},\epsilon_{\rm num}\right)}.
\label{eq:physics_reward}
\end{equation}
Here $\lambda$ sets reward strength, $\chi_t=b_tc_t$ weights label reliability, and $\rho(i)$ introduces the reward with a half-cosine ramp.
The event mask $m_{k,t}$ activates term $k$, weighted by $\beta_k(i)$; $\epsilon_{\rm num}>0$ prevents division by zero.
In Table~\ref{tab:training_parameters}'s $\boldsymbol\beta$ order, $\mathcal K$ comprises contact schedule, vertical-load share, total vertical force, per-foot force, touchdown impulse, and signed contact change.
These terms compare contact throughout the gait, load/force during stable support, impulses around touchdown, and signed changes for Canter touchdown/liftoff.
Each $R_{k,t}=\exp(-e_{k,t}/\sigma_k)$ uses normalized mean squared error $e_{k,t}$ and tolerance $\sigma_k>0$.
Force normalization averages forward- and backward-bank medians of total vertical force over loaded frames; impulse normalization analogously averages median touchdown-impulse magnitudes.
Contact probabilities and load shares are dimensionless.
Only the signed-change weight $\beta_{\rm change}(i)$ is ramped, using a half-cosine schedule; Table~\ref{tab:training_parameters} gives final weights and schedule intervals.
After its half-cosine decay, $\alpha(i)$ stays zero; $\rho(i)$ is zero before its half-cosine ramp and one afterward.
The physics reward augments style reward without changing task rewards.

\subsection{Demonstration-guided Omnidirectional Command Expansion (DOCE)}
To avoid forcing distant command tracking before learning the demonstrated gait, DOCE starts from frame-matched reference velocities and progressively expands to the target command ranges.
Each resampling step selects the final reference condition, then samples around its current frame:
\begin{equation}
\begin{aligned}
\tilde{\mathbf c}={}&(1-p)\mathbf c^E_{s_x,s_r}
+p\,S_{s_x,s_r}(\mathbf c_{\max}\odot\boldsymbol\xi),\\
\mathbf c={}&\operatorname{clip}(\tilde{\mathbf c},-\mathbf c_{\max},\mathbf c_{\max}),
\qquad v_x\leftarrow s_x|v_x|.
\end{aligned}
\label{eq:command_expansion}
\end{equation}
Here $\mathbf c=(v_x,v_y,\omega_z)$, $\mathbf c_{\max}$ contains the three target bounds, and the components of $\boldsymbol\xi$ are independently sampled from $\mathcal U(-1,1)$.
The signs $s_x,s_r\in\{-1,+1\}$ select the forward/time-reversed and original/sagittally reflected reference, respectively; $S_{s_x,s_r}=\operatorname{diag}(s_x,s_r,s_r)$ applies the same geometric transformation to the paired random draw.
Crucially, $\mathbf c^E_{s_x,s_r}$ is read from the exact frame of this already selected reference, rather than from a gait-wide mean or a reference chosen afterward.
The same $(s_x,s_r)$ condition selects the kinematic and physical targets used by DMA.
We balance each gait's four reference conditions within local cycle-progress bins to keep motion stages comparable.

The progress $p=\min(i/I_c,1)$ moves the sampling center from the selected reference toward a full-range random command over $I_c$ iterations; these ranges are retained for the final $I_{\rm full}$ iterations.
Because $\xi_y$ and $\xi_\omega$ are independent, lateral and yaw commands cover all four sign combinations rather than sharing a command sign.
With balanced $s_x$, the final curriculum spans the full Cartesian product of the target per-axis ranges in Fig.~\ref{fig:method_overview}(b).

\subsection{Shared-residual Multi-gait Policy (SMP)}
SMP shares balance and velocity control while retaining gait-specialized corrections.
A shared actor receives state, command, and a four-dimensional one-hot gait code; four residual branches ($K=4$) receive state and command in parallel, without the gait code.
A gait-conditioned soft gate combines their corrections with the complete shared action:
\begin{equation}
\mathbf a=\mathbf a_{\mathrm{sh}}(\mathbf o,\mathbf z)
+\gamma_r\sum_{k=1}^{K}w_k(\mathbf z)\,\mathbf r_k(\mathbf o).
\label{eq:shared_residual_actor}
\end{equation}
Here $\mathbf o$ contains state and command, $\mathbf z$ is the gait code, and $\mathbf a_{\rm sh}$ is the shared action.
Softmax weights $w_k$ sum to one, $\mathbf r_k$ are per-joint $\tanh$-bounded residuals, and $\gamma_r$ scales their mixture.
Sharing refers to parameters across gaits, not gait-independent actions.
The gate's learned linear map initially favors a different branch per gait but permits soft reuse; the branches are residual corrections, not independent gait policies.
The shared path supplies a complete action even when residual corrections are small, while soft mixing allows each gait to reuse corrections learned by other branches.

State comprises body angular velocity, projected gravity, joint positions/velocities, and previous action; policy outputs specify scaled joint-position offsets for PD control.
The actor parameter count $|\theta_\pi|$ matches the dense baseline.
All branches train jointly from scratch with PPO~\cite{schulman2017ppo}, using separate style and task value estimates and averaging their normalized advantages~\cite{sood2025apex}; completed labels are not critic inputs.
Evaluation and deployment use only the actor, proprioception, command, and gait code, without references, reference-frame indices, cycle progress, physical labels, critics, or torque guidance.
\looseness=-1

\section{Results}

Our experiments address three questions:
(1) How does OmniMimic compare with AMP and APEX under a common evaluation protocol for forward and backward motions?
(2) How do command expansion, the shared-residual policy, and dynamics-completed supervision affect motion fidelity and command tracking?
(3) Does constrained dynamics completion provide more effective backward supervision than missing or naively reversed physical labels?

\subsection{Experimental Setup}

We evaluated \emph{Trot}, \emph{Pace}, \emph{Canter}, and \emph{Pronk} on a Unitree Go2 in Isaac Gym~\cite{makoviychuk2021isaacgym}.
Each method learned a single policy for all gaits.
Each forward motion was paired with its strict temporal reversal (Sec.~\ref{sec:method}); OmniMimic and its data-generation controls additionally applied sagittal reflection to both kinematic and physical targets.
All methods used the same source motions, command bounds, and task rewards.
Table~\ref{tab:training_parameters} summarizes the numerical settings.

\begin{table}[!t]
    \centering
    \caption{Training and implementation parameters.}
    \label{tab:training_parameters}
    \vskip-1ex
    \footnotesize
    \setlength{\tabcolsep}{3pt}
    \renewcommand{\arraystretch}{1.04}
    \begin{tabular*}{\columnwidth}{@{}l@{\extracolsep{\fill}}ll@{}}
        \toprule
        Group & Symbol & Value \\
        \midrule
        \multirow[t]{8}{*}{\textbf{DMA}}
          & $d_t$ & $0.998^{n_t/100}$ \\
          & $(w_v,w_\tau,w_s,w_f)$ & $(0.12,0.016,0.004,0.03)$ \\
          & $\gamma_\tau$ & $0.05$ \\
          & $\alpha(i)$ & $1\!\rightarrow\!0,\ i\in[0,2{,}000]$ \\
          & $\boldsymbol\beta$ & $(0.55,0.08,0.04,0.08,0.25,0.02)$ \\
          & $\beta_{\rm change}(i)$ (Canter) & $0\!\rightarrow\!0.02,\ i\in[1{,}800,2{,}200]$ \\
          & $\lambda$ & $0.2925$ \\
          & $\rho(i)$ & $0\!\rightarrow\!1,\ i\in[700,2{,}000]$ \\
        \midrule
        \multirow[t]{4}{*}{\textbf{DOCE}}
          & $c_{x,\max},c_{y,\max}$ & $2\,\mathrm{m/s}$ \\
          & $c_{\omega,\max}$ & $1.5\,\mathrm{rad/s}$ \\
          & $I_c$ & $2{,}000$ \\
          & $I_{\rm full}$ & $400$ \\
        \midrule
        \multirow[t]{2}{*}{\textbf{SMP}}
          & $\gamma_r$ & $0.2$ \\
          & $|\theta_\pi|$ & $191{,}372$ \\
        \midrule
        \multirow[t]{4}{*}{\textbf{Training}}
          & $N_{\rm env}$ & $4{,}096$ \\
          & $(w_{\rm style},w_{\rm task})$ & $(0.5,0.5)$ \\
          & $I_{\rm Ours}=I_{\rm APEX}$ & $2{,}400$ \\
          & $I_{\rm AMP}$ & $10{,}000$ \\
        \bottomrule
    \end{tabular*}
    \par\smallskip
    \begin{minipage}{\columnwidth}
    \footnotesize
\textbf{DMA implementation.}
OSQP settings: polishing, absolute/relative tolerance $10^{-7}$, iteration limit $10^5$.
The $42$-variable QP used inverse-squared scales $(10,30,80)$ for base translation/rotation/joint accelerations (SI units) and $(\overline\tau_{\max},mg)$ for torque/force; smoothing used $0.2$ times these physical weights after the first frame.
Here $\overline\tau_{\max}$ is the mean URDF effort limit; $m$ is robot mass, $\mu=1$ and $f_{z,\max}=4mg$.
Finite labels required dynamics-residual norm $\leq10^{-4}$ and actuator/contact violations $\leq2\!\times\!10^{-4}$.
Failed solves aborted generation; invalid labels were masked during training.
    \end{minipage}
    \vskip-4ex
\end{table}

\subsubsection{Evaluation Metrics}

All simulation evaluations used the actor alone, with torque guidance and domain randomization disabled.

\emph{Reference-aligned forward and backward tracking.}
For each gait and direction, we ran $128$ trajectories initialized at evenly spaced reference frames for one reference-clip duration (approximately $8$\,s), accumulating errors only before the first episode termination. Forward and backward runs used the original and time-reversed references, respectively; each command $(v_x,v_y,\omega_z)$ came from the corresponding reference frame.
The metrics $q$, $h$, and $x_{ee}$ are reference-frame RMSEs in the $12$ joint angles (rad), base height (m), and four-foot positions (m); foot positions are base-relative and yaw-aligned. Cmd. pools the errors in body-frame $(v_x,v_y,\omega_z)$ into an RMSE without per-axis normalization, mixing m/s, m/s, and rad/s. We average directional RMSEs equally within each gait, then equally across gaits.

Force (N) and Load (unitless) measure yaw-aligned contact-force and vertical-load-share RMSE on reference-designated stable support feet. TD (N\,s) measures three-dimensional contact-impulse RMSE over valid $0.1$\,s touchdown windows. All variants use the same simulation-derived forward and completed backward labels, averaged equally over directions and gaits.

\begin{figure}[!t]
    \centering
    \includegraphics[width=\columnwidth]{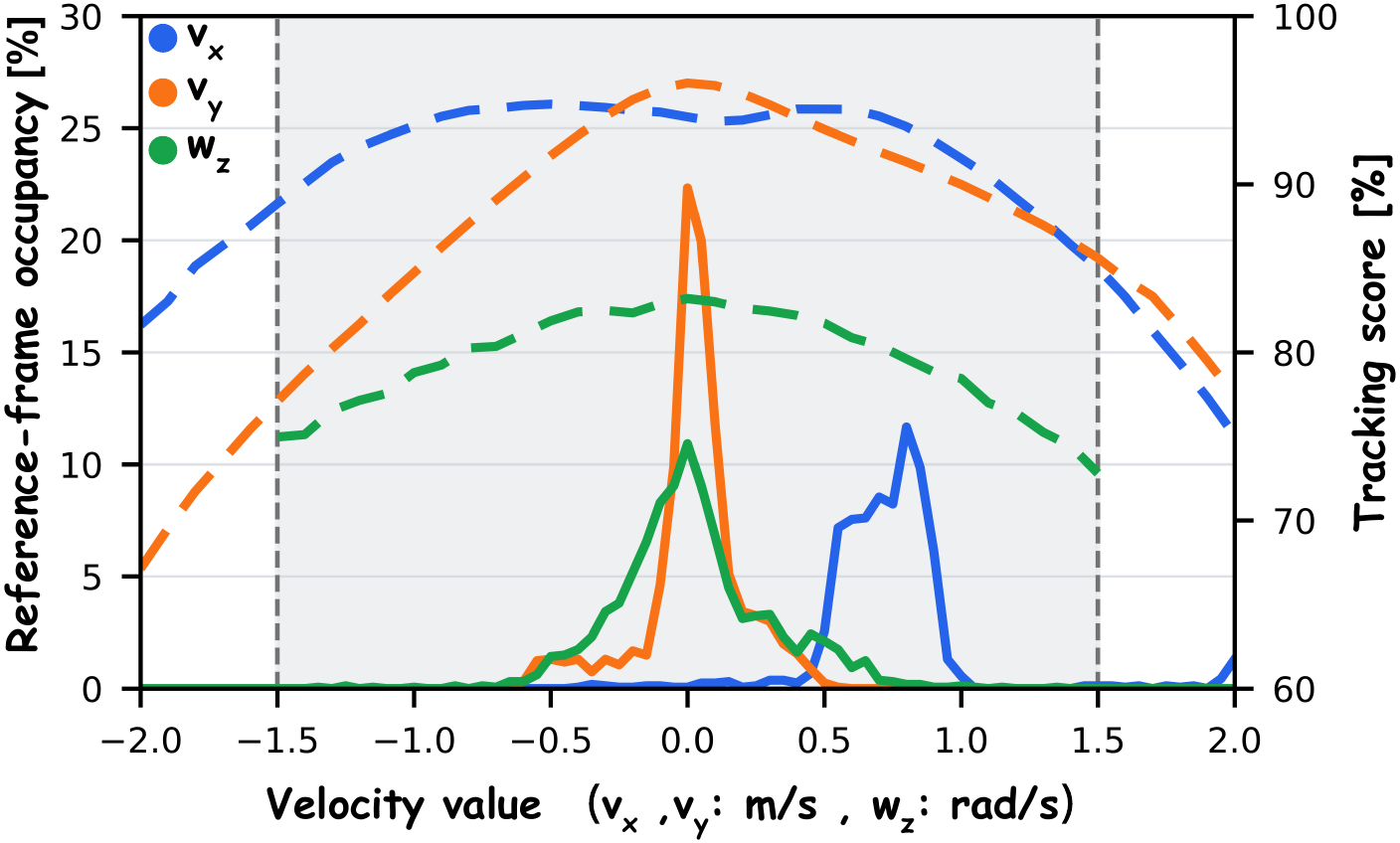}
    \vskip-1ex
    \caption{Reference-motion velocity distributions (solid) and OmniMimic's marginal tracking agreement (dashed). Dashed curves use RMSE pooled across four gaits and averaged over the other two command axes on the $52{,}111$-command grid.}
    \label{fig:command_distribution_coverage}
    \vskip-3ex
\end{figure}

\emph{Dense Cartesian command grid.}
The tabulated Grid-$9^3$ averages the same three-component RMSE over $729$ fixed commands and four gaits, independently of reference velocities. Each axis has nine uniformly spaced values, with $v_x,v_y\in[-2,2]$\,m/s and $\omega_z\in[-1.5,1.5]$\,rad/s. Each command receives one rollout of $1{,}602$ control steps: the first $50$ steps ($1$\,s) are discarded, and terminal velocity is held for the remaining measurement window after early termination.

Figs.~\ref{fig:command_distribution_coverage} and~\ref{fig:velocity_radar} use a finer $41\!\times\!41\!\times\!31$ grid ($52{,}111$ commands per gait) with $0.1$ increments over the same bounds and the same termination protocol. Their percentage agreement is one minus the aggregated axis RMSE divided by that axis's maximum absolute command, clipped to $[0,1]$ and multiplied by $100$; it is not a success rate. Per-gait scores average axis RMSEs over commands; the all-gait aggregate pools squared errors equally across gaits before taking the root and averaging over commands.
Fig.~\ref{fig:command_distribution_coverage} shows that command tracking extends beyond the velocity regions most densely represented in the reference motions.

\subsection{Comparison with AMP and APEX}

\textbf{Baselines.}
AMP~\cite{peng2021amp} and APEX~\cite{sood2025apex} used dense actors and sampled command components independently and uniformly over the fixed bounds, without DOCE. APEX used action-prior training and shared OmniMimic's advantage weighting and training budget. AMP retained its mixed-reward objective without the torque guidance used by APEX and OmniMimic; we allocated it a longer budget to allow additional convergence time.

\begin{table}[!t]
    \centering
    \caption{Comparison of AMP~\cite{peng2021amp}, APEX~\cite{sood2025apex}, and \mbox{OmniMimic}. Mean averages four gaits; reference-aligned metrics also weight forward and backward directions equally. Lower is better; best in bold.
    }
    \label{tab:independent_baselines}
    \vskip-1ex
    \scriptsize
    \setlength{\tabcolsep}{3.0pt}
    \resizebox{\columnwidth}{!}{%
    \begin{tabular}{l l c c c c c}
        \toprule
        Motion & Method & $q\downarrow$ & $h\downarrow$ & $x_{ee}\downarrow$
        & Cmd.$\downarrow$ & Grid-$9^3\downarrow$ \\
        \midrule
        Trot & AMP & \textbf{0.2530} & 0.0257 & 0.0625
          & 0.4175 & 1.0323 \\
        & APEX  & 0.2637 & \textbf{0.0229}
          & 0.0634 & \textbf{0.2021} & 0.7711 \\
        & Ours  & 0.2681 & 0.0270 & \textbf{0.0624} & 0.2140
          & \textbf{0.1928} \\
        \cmidrule(lr){1-7}
        Pace & AMP  & 0.2624 & 0.0274 & 0.0629
          & 0.3674 & 0.9804 \\
        & APEX  & 0.2091 & \textbf{0.0221} & 0.0497
          & \textbf{0.2494} & 0.8210 \\
        & Ours  & \textbf{0.2041} & 0.0243 & \textbf{0.0457}
          & 0.2549 & \textbf{0.2009} \\
        \cmidrule(lr){1-7}
        Canter & AMP  & \textbf{0.3793} & \textbf{0.0260} & 0.0979
          & 0.4371 & 1.1748 \\
        & APEX & 0.4214 & 0.0425
          & 0.0930 & 0.3588 & 0.7061 \\
        & Ours  & 0.4017 & 0.0473 & \textbf{0.0833} & \textbf{0.2647}
          & \textbf{0.2844} \\
        \cmidrule(lr){1-7}
        Pronk & AMP  & 0.2534 & 0.0774 & 0.0558
          & 0.2533 & 1.0079 \\
        & APEX  & 0.2423 & 0.0702 & 0.0567
          & \textbf{0.1664} & 0.8273 \\
        & Ours  & \textbf{0.1769} & \textbf{0.0443}
          & \textbf{0.0374} & 0.1729 & \textbf{0.4738} \\
        \midrule
        Mean & AMP  & 0.2870 & 0.0391 & 0.0698
          & 0.3688 & 1.0488 \\
        & APEX & 0.2841 & 0.0394 & 0.0657
          & 0.2442 & 0.7814 \\
        & \textbf{Ours} & \textbf{0.2627} & \textbf{0.0357} & \textbf{0.0572}
          & \textbf{0.2266} & \textbf{0.2880} \\
        \bottomrule
    \end{tabular}
    }
    \vskip-2ex
\end{table}

In Table~\ref{tab:independent_baselines}, OmniMimic achieved lower four-gait mean errors than both baselines on all five metrics. Relative to APEX, foot-position RMSE decreased by $12.9\%$ and grid command RMSE by $63.1\%$. These gains reflect the complete framework; Table~\ref{tab:algorithm_ablation} separates the components' effects on motion fidelity and command tracking.

\begin{figure}[!t]
    \centering
    \includegraphics[width=\columnwidth]
{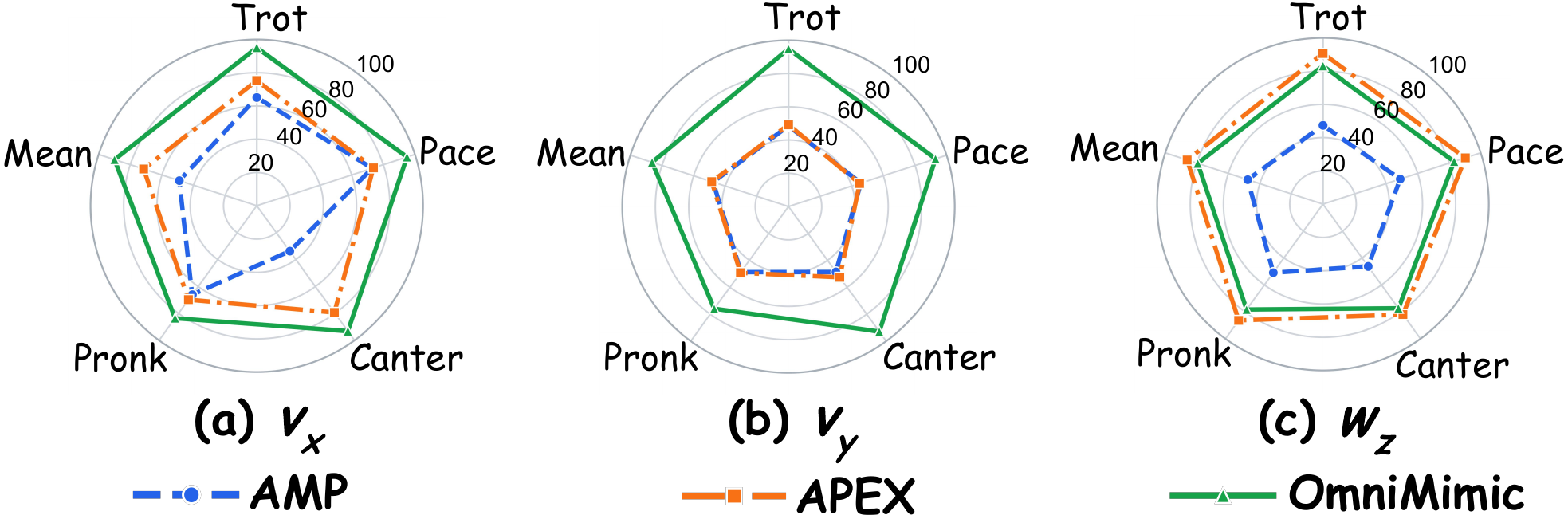}
    \caption{Velocity-tracking agreement for (a) $v_x$, (b) $v_y$, and (c) $\omega_z$. Each panel compares AMP~\cite{peng2021amp}, APEX~\cite{sood2025apex}, and OmniMimic across four gaits and the all-gait aggregate, using $52{,}111$ deterministic commands per gait. Agreement is the normalized RMSE-based score defined in Experimental Setup; higher is better.}
    \label{fig:velocity_radar}
    \vskip-3ex
\end{figure}

\begin{figure*}[!t]
    \centering
    \includegraphics[width=\textwidth]{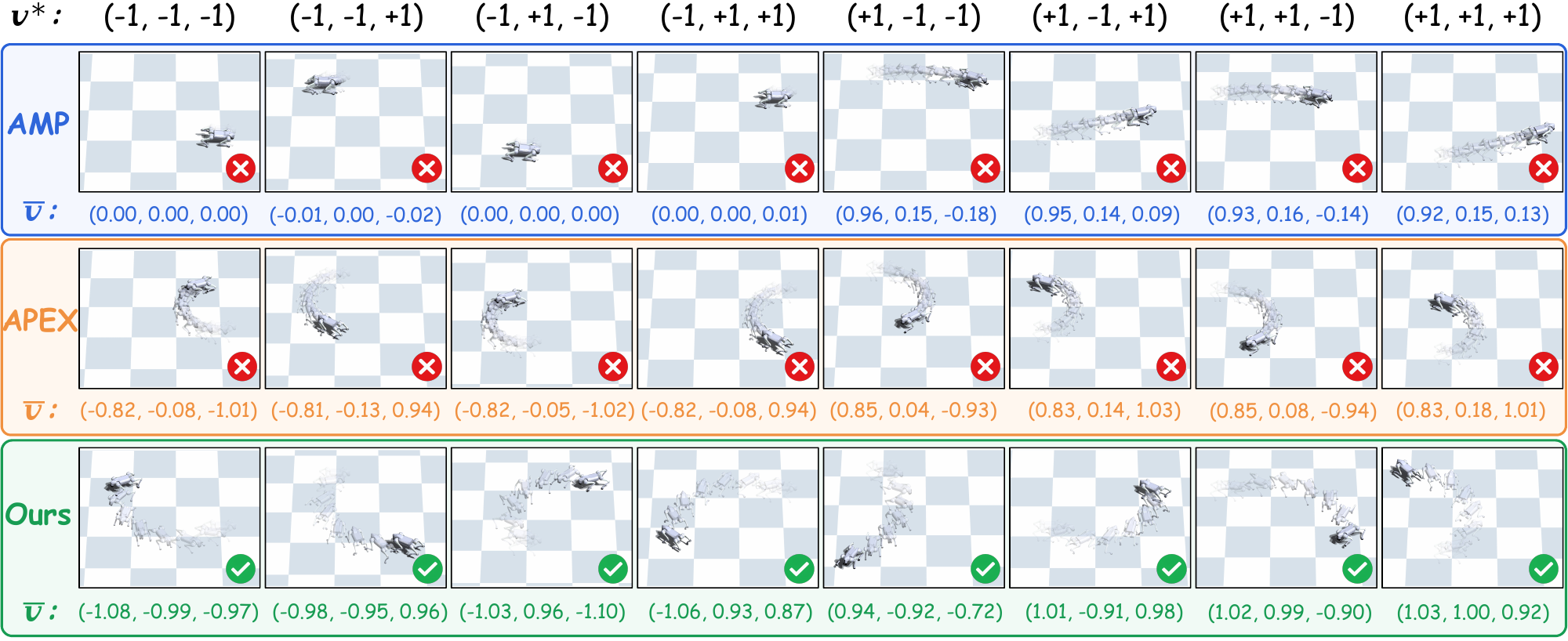}
    \caption{Trot rollouts in simulation for AMP~\cite{peng2021amp}, APEX~\cite{sood2025apex}, and OmniMimic (top to bottom), from the same standing pose without torque guidance, across all eight sign combinations of the command $\mathbf v^\star=(v_x^\star,v_y^\star,\omega_z^\star)$ (columns); translational components are $\pm1$\,m/s and yaw is $\pm1$\,rad/s. Each panel shows the opaque $2.8$\,s pose, progressively more transparent earlier poses, and mean body-frame velocities $\bar{\mathbf v}=(\bar v_x,\bar v_y,\bar\omega_z)$ over $1$--$6$\,s; green checks/red crosses mark qualitative success/failure in executing the combined command.}
    \label{fig:trot_qualitative}
    \vskip-3ex

\end{figure*}

Fig.~\ref{fig:velocity_radar} shows the largest gains in translation: \mbox{OmniMimic} reached $90.0\%$ and $86.2\%$ all-gait agreement for $v_x$ and $v_y$, exceeding APEX by $18.4$ and $37.7$ percentage points. APEX scored higher in yaw ($86.1\%$ versus $79.4\%$).

In Fig.~\ref{fig:trot_qualitative}, OmniMimic followed all eight combined translational and yaw commands.
APEX tracked yaw closely but showed weak lateral motion.
AMP remained nearly stationary for negative longitudinal commands and moved mainly forward for positive ones.

\subsection{Learning-module Ablation}
Table~\ref{tab:algorithm_ablation} separates the contributions to command tracking and motion fidelity through a progressive ablation.
Starting from a dense policy with uniform command sampling (A), B introduces DOCE, C replaces the actor with parameter-matched SMP, and D adds DMA.
All other training and evaluation conditions are held fixed.

\begin{table}[!t]
    \centering
    \caption{Learning-module ablation. $\checkmark$/$\times$ denote included/excluded components. Results use the same aggregation as Table~\ref{tab:independent_baselines}. Lower is better; best in bold.}
    \label{tab:algorithm_setting}
    \label{tab:algorithm_ablation}
    \vskip-1ex
    \footnotesize
    \setlength{\tabcolsep}{2.0pt}
    \renewcommand{\arraystretch}{1.12}
    \resizebox{\columnwidth}{!}{%
    \begin{tabular}{c c c c c c c c c}
        \toprule
        \multirow{2}{*}{Var.} & \multicolumn{3}{c}{Configuration}
          & \multicolumn{5}{c}{Four-gait mean results} \\
        \cmidrule(lr){2-4}\cmidrule(lr){5-9}
        & DOCE & SMP & DMA
          & $q\downarrow$ & $h\downarrow$ & $x_{ee}\downarrow$
          & Cmd.$\downarrow$ & Grid-$9^3\downarrow$ \\
        \midrule
        A & $\times$ & $\times$ & $\times$
          & 0.2841 & 0.0394 & 0.0657 & 0.2442 & 0.7814 \\
        B & $\checkmark$ & $\times$ & $\times$
          & 0.2893 & 0.0381 & 0.0653 & 0.2265 & 0.3406 \\
        C & $\checkmark$ & $\checkmark$ & $\times$
          & 0.2879 & 0.0366 & 0.0645 & \textbf{0.2163} & \textbf{0.2866} \\
        D & $\checkmark$ & $\checkmark$ & $\checkmark$
          & \textbf{0.2627} & \textbf{0.0357} & \textbf{0.0572} & 0.2266 & 0.2880 \\
        \bottomrule
    \end{tabular}%
    }
    \vskip-3ex
\end{table}

DOCE (A to B) supplied the largest grid-tracking gain, reducing RMSE by $56.4\%$, although joint-angle RMSE increased by $1.8\%$. SMP (B to C) further reduced grid RMSE by $15.9\%$ and improved the other four metrics.
Adding DMA (C to D) improved all three motion-reproduction metrics, reducing foot-position RMSE from $0.0645$ to $0.0572$ ($11.3\%$), while grid RMSE changed from $0.2866$ to $0.2880$ and reference-command error also increased slightly.
These comparisons suggest complementary roles: DOCE and SMP primarily improve command tracking, whereas DMA improves motion fidelity at reference-aligned velocities.

\subsection{Data-generation Ablation}
\label{sec:data_ablation}
To isolate dynamics completion, we varied only backward physical supervision, retaining the same forward targets, time-reversed and reflected kinematics, and learning pipeline:
\begin{itemize}
    \item \textbf{R1}: backward physical supervision is masked, disabling the physical residual guide and physics reward;
    \item \textbf{R2}: force, impulse, and torque labels obtained from the offline tracking policy's forward rollouts are directly time-reversed before sagittal reflection;
    \item \textbf{R3}: constrained dynamics completion reconstructs backward labels before the same sagittal reflection.
\end{itemize}
R3 and variant D (Table~\ref{tab:algorithm_ablation}) use the same policy.

\begin{table}[!t]
    \centering
    \caption{Backward physical-label ablation, evaluated with common reference labels and the same aggregation as Table~\ref{tab:independent_baselines}. Force, Load, and TD are contact-force (N), load-share (unitless), and touchdown-impulse (N\,s) RMSE. Lower is better; best in bold.}
    \label{tab:data_ablation}
    \vskip-1ex
    \scriptsize
    \setlength{\tabcolsep}{3.0pt}
    \resizebox{\columnwidth}{!}{%
    \begin{tabular}{c c c c c c c c c}
        \toprule
        Var. & $q\downarrow$ & $h\downarrow$ & $x_{ee}\downarrow$
        & Cmd.$\downarrow$ & Grid-$9^3\downarrow$
        & Force$\downarrow$ & Load$\downarrow$ & TD$\downarrow$ \\
        \midrule
        R1 & 0.2832 & 0.0407 & 0.0641 & 0.2229 & 0.3286
           & 34.584 & 0.3292 & 2.504 \\
        R2 & 0.2816 & 0.0451 & 0.0643 & \textbf{0.2229} & \textbf{0.2865}
           & 35.158 & 0.3391 & 2.440 \\
        \textbf{R3} & \textbf{0.2627} & \textbf{0.0357}
           & \textbf{0.0572} & 0.2266 & 0.2880
           & \textbf{33.202} & \textbf{0.3245} & \textbf{2.267} \\
        \bottomrule
    \end{tabular}%
    }
    \vskip-3ex
\end{table}

R3 achieved the lowest error on six of eight metrics (Table~\ref{tab:data_ablation}).
Compared with naive reversal (R2), completion reduced contact-force and touchdown-impulse RMSE by $5.6\%$ and $7.1\%$, while reference-command and dense-grid errors were $1.7\%$ and $0.5\%$ higher.
R3 also improved seven of eight metrics over missing backward supervision (R1). Under the common evaluation targets, completion improved motion fidelity and agreement with the physical reference labels, with a small command-tracking trade-off.

\subsection{Real-world Deployment}
\begin{figure}[!t]
    \centering
    \includegraphics[width=\columnwidth]{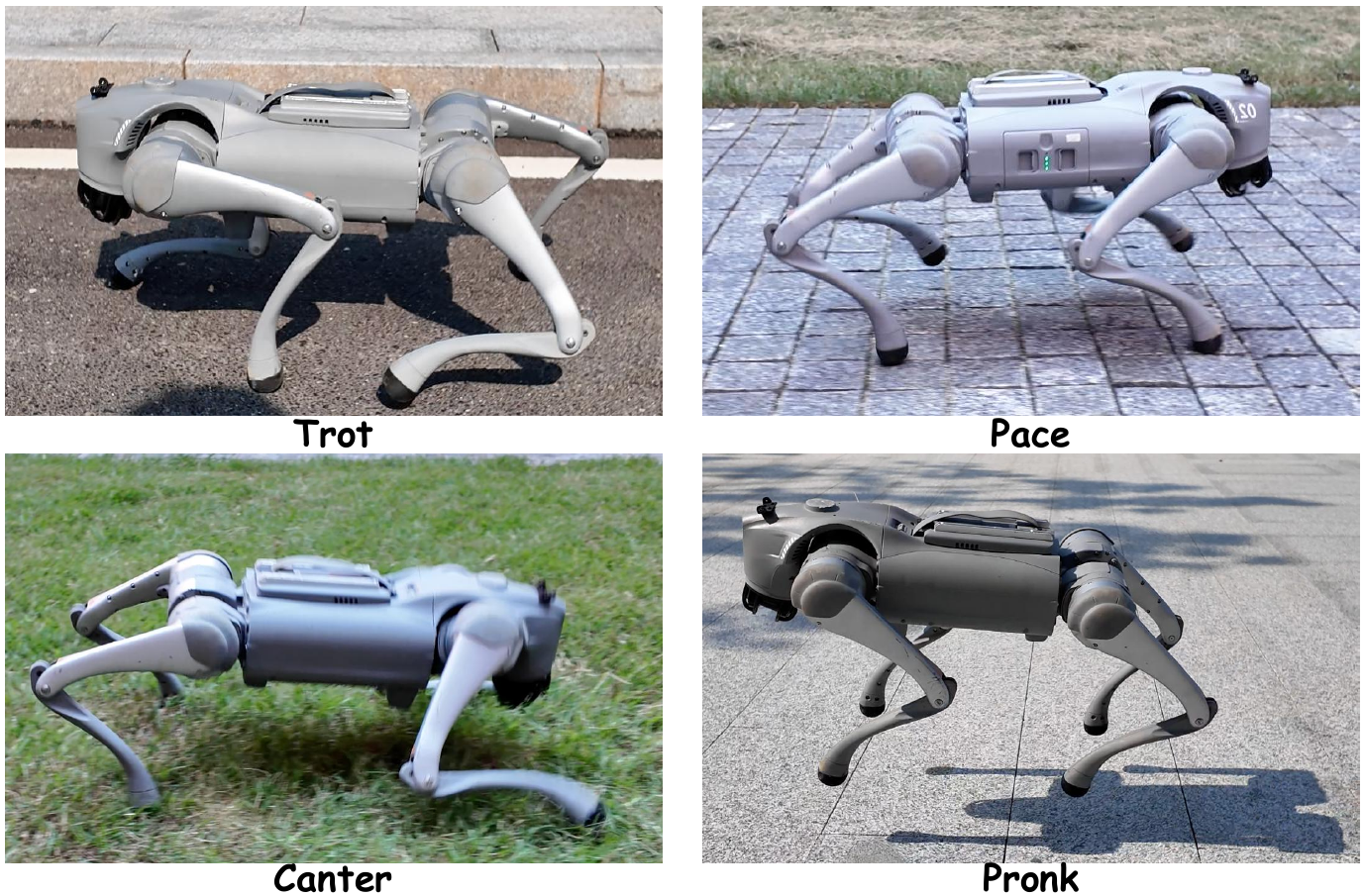}
    \vskip-1ex
    \caption{Hardware demonstrations. A single simulation-trained OmniMimic policy executes \emph{Trot}, \emph{Pace}, \emph{Canter}, and \emph{Pronk} on the Unitree Go2 without fine-tuning.}
    \label{fig:terrain_results}
    \vskip-3ex
\end{figure}

We deployed the simulation-trained OmniMimic policy directly on the Unitree Go2 without fine-tuning.
The actor received proprioceptive observations, velocity commands, and a gait code, and produced joint-position targets for PD control.
Reference motions, completed physical labels, and guidance torques were not required during execution.
Fig.~\ref{fig:terrain_results} shows \emph{Trot}, \emph{Pace}, and \emph{Pronk} on paved surfaces and \emph{Canter} on grass. 
The project-page videos further demonstrate forward, backward, lateral, and turning motions on the physical robot.

\section{Conclusion}
We presented OmniMimic, which combines dynamics-completed motion augmentation, progressive command expansion, and a shared-residual policy to learn multi-gait omnidirectional control from directionally limited animal demonstrations.

Simulation results show improvements in reference-motion fidelity and command tracking over the evaluated baselines. Ablations suggest complementary roles: command expansion and the shared-residual policy primarily improve tracking, while dynamics-completed supervision improves motion fidelity at reference-aligned velocities with a small tracking trade-off. All four gaits were demonstrated on the Unitree Go2 without fine-tuning or deployment-time reference inputs and guidance torques.

Motion fidelity is quantified only at reference-aligned forward and backward velocities. The current formulation also uses discrete gait labels and is primarily evaluated on flat terrain. \textbf{Future work} will examine gait coordination across commands and quantify hardware tracking accuracy and repeatability over broader command and terrain ranges.

\bibliographystyle{IEEEtran}
\bibliography{bib}

\end{document}